\documentclass[10pt,twocolumn,letterpaper]{article}

\usepackage{cvpr}              
\usepackage{amsmath}
\usepackage{amssymb}
\usepackage{multirow}
\usepackage{booktabs}

\usepackage[dvipsnames]{xcolor}

\definecolor{cvprblue}{rgb}{0.21,0.49,0.74}
\usepackage[pagebackref,breaklinks,colorlinks,citecolor=cvprblue]{hyperref}

\def\paperID{} 
\def\confName{CVPR}
\def\confYear{2024}

\title{Text-IF: Leveraging Semantic Text Guidance for Degradation-Aware and Interactive Image Fusion}

\author{Xunpeng Yi, Han Xu, Hao Zhang, Linfeng Tang, Jiayi Ma\thanks{Corresponding author}\\
Electronic Information School, Wuhan University, Wuhan 430072, China\qquad\\
{\tt\small \{yixunpeng, xu\_han\}@whu.edu.cn,\,\{zhpersonalbox, linfeng0419, jyma2010\}@gmail.com}}

\begin{document}
\maketitle

\begin{abstract}
Image fusion aims to combine information from different source images to create a comprehensively representative image. Existing fusion methods are typically helpless in dealing with degradations in low-quality source images and non-interactive to multiple subjective and objective needs. To solve them, we introduce a novel approach that leverages semantic text guidance image fusion model for degradation-aware and interactive image fusion task, termed as Text-IF. It innovatively extends the classical image fusion to the text guided image fusion along with the ability to harmoniously address the degradation and interaction issues during fusion. Through the text semantic encoder and semantic interaction fusion decoder, Text-IF is accessible to the all-in-one infrared and visible image degradation-aware processing and the interactive flexible fusion outcomes. In this way, Text-IF achieves not only multi-modal image fusion, but also multi-modal information fusion. Extensive experiments prove that our proposed text guided image fusion strategy has obvious advantages over SOTA methods in the image fusion performance and degradation treatment. The code is available at https://github.com/XunpengYi/Text-IF.
\end{abstract}

\section{Introduction}
\label{sec:intro}
Image fusion is a prominent field within the domain of digital image processing~\cite{liu2023multi, xu2022rfnet, sun2022detfusion}. Single-modal images can only capture partial representation of the scene. Multi-modal images allow for the effective acquisition of more comprehensive representation. As an important representative, visible images provide the reflectance-based visual information, akin to human vision. Infrared images provide thermal radiation-based information, more valuable for detecting thermal targets and observing nighttime activities. The infrared and visible image fusion focuses on fusing the complementary information of infrared and visible images, yielding high-quality fusion images~\cite{ma2019infrared,ma2020ganmcc,ma2020ddcgan,zhang2021image,
zhang2021sdnet,zhu2022clf,tang2022image}.

\begin{figure}[!t]
\centering
\includegraphics[width=0.97\linewidth]{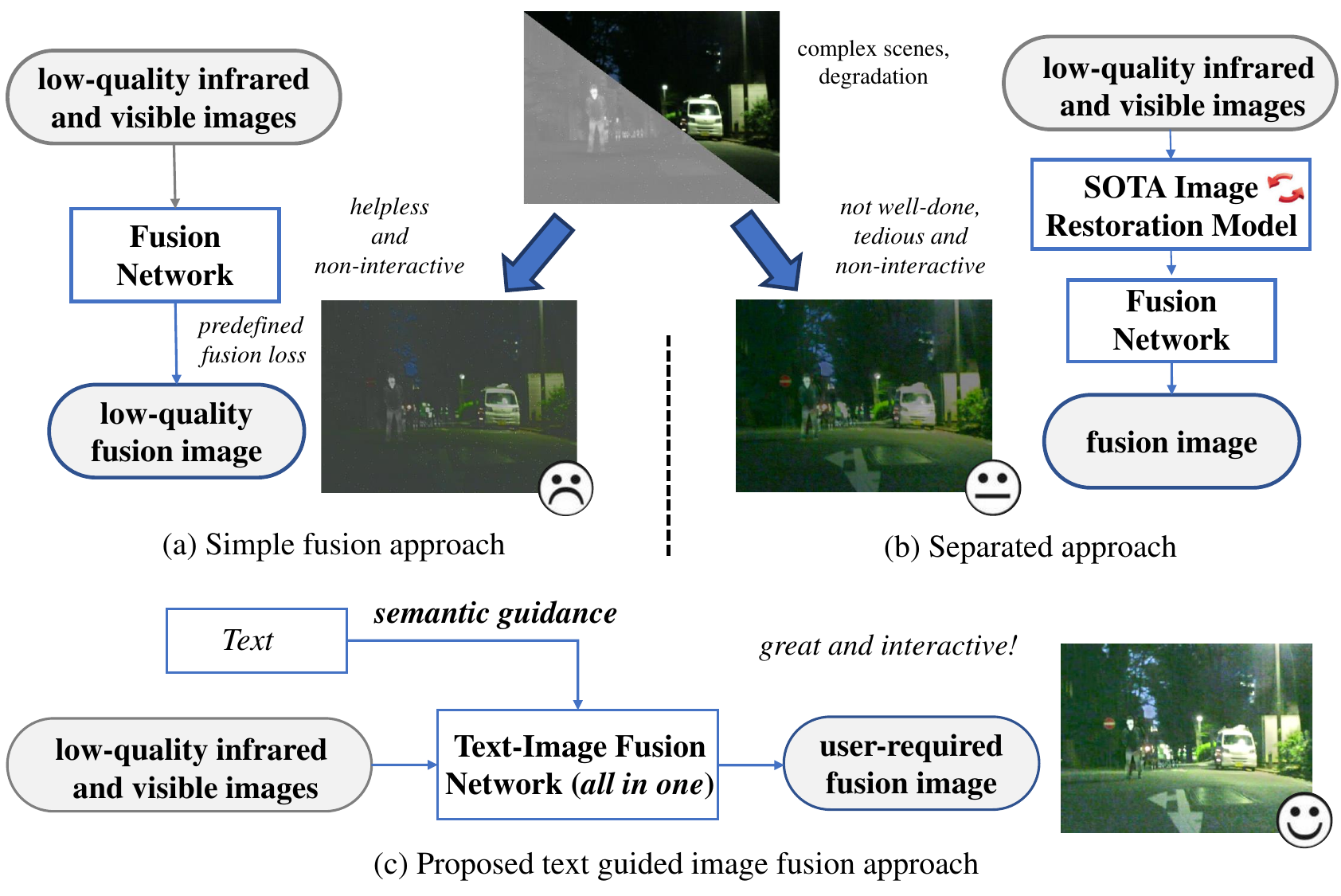}
\vspace{-0.1in}
\caption{Fusion approaches for complex scenes with degradations. (a) simple fusion approach: treating image fusion with predefined fusion loss and not applicable to complex scenes with degradations. (b) separated approach: requiring frequent restoration methods switching according to the type of degradations, which is troublesome and not well-done. (c) proposed text guided image fusion approach: achieving interactive and high-quality fusion image without tedious replacement of models.}
\label{proposed framekwork}
\end{figure}

Limited by the conditions of environments, the originally acquired infrared and visible images may suffer from degradations and show low fusion image quality. The visible images are susceptible to degradation issues, \textit{e.g.}, low light, over exposure, \textit{etc}. The infrared images are inevitably affected by noise (including thermal, electronic, and environmental noise), diminished contrast, and other associated effects. Current fusion methods lack the capability to adaptively solve the degradations, leading to the low-quality fusion image. Furthermore, relying on manual pre-processing to enhance the image has the problems of flexibility and efficiency~\cite{tang2022piafusion}. Therefore, it is of practical interest to study a model that harmonises degradation-aware processing and interactive fusion.

Designing a model for individualized degradation to achieve image enhancement and fusion is feasible. However, most of image fusion tasks need to be carried out in various complex conditions around the clock. As shown in Fig.~\ref{proposed framekwork}, it means that multiple image restoration models are needed to match the requirements, which requires frequent models switching and brings great consumption and trouble. In addition, the separation approach has the problem to achieve harmony between the enhancement and fusion, resulting in unsatisfactory overall performance.

In addition, the real-world image fusion is complex, flexible and task-oriented. The requirements of the image fusion may change according to the subjective needs of users and objective application tasks. In all scenarios, if the method is non-interactive and produces a relatively fixed fusion result, it usually falls short for various and flexible requirements of the users.

As an important way of human and machine interaction, text is widely used in the model of specifying requirements. Recent research in large-scale visual language has achieved amazing results in image generation~\cite{rombach2022high, liu2022compositional, liao2022text, tao2022df}, demonstrating the potential of this paradigm. The interaction between semantic text and image processing procedures can achieve the goal of customized image processing. In addition, PromptIR~\cite{potlapalli2023promptir} proposed learnable visual prompts and implemented various degradation removals, but not realizes text guided and lacks the design for multi-modal degradations and fusion. Therefore, it is of great significance to implement the degradation-aware processing and user interactivity in image fusion by text.

To this end, we propose a model that leverages the semantic text guidance for degradation-aware and interactive image fusion, termed as Text-IF. It integrates the text and image fusion to meet the needs of harmonious degradation-aware processing and interactivity fusion. Especially, it allows the text to provide the flexible semantic guidance to deal with various degradations, which is a type of multi-modal information fusion. In general, Text-IF contains the image pipeline, and text interaction guidance architecture, including the text semantic encoder and the semantic interaction guidance module. In the image fusion pipeline, we meticulously design the Transformer-based image extraction module and the cross fusion layer for high-quality fusion. In the text semantic encoder, we aggregate the text semantic extraction capabilities of powerful pre-trained vision-language models. Through the semantic interaction guidance module, the semantic features of text and image fusion features are coupled together to achieve the goal of text guided image fusion. It solves the problem that the existing image fusion methods are difficult to adapt to the fusion of complex scenes with degradations, and can only output relatively fixed results without interactivity. It provides a feasible direction for the subsequent research of text guided image fusion tasks.

Overall, our contributions can be summarized as follows:
\begin{itemize}
    \item To adapt complex degradation conditions, we address the integrated problem of image fusion and degradation-aware processing. It breaks through the limitation of quality improvement in image fusion.
    \item We introduce a semantic interaction guidance module to fuse the information of text and images. The proposed method achieves not only multi-modal image fusion, but also multi-modal information fusion. 
    \item The proposed method ultimately increases freedom of customized fusion results. It provides the interactive fusion and can generate more flexible, high-quality and user-required results without prior expertise or predefined rules.
\end{itemize}

\section{Related Work}

\textbf{General Image Fusion Methods.}
General Image fusion methods have made significant advancements with the advent of deep learning. During the early phase, fusion strategies based on pre-trained autoencoders were extensively employed. CSR~\cite{liu2016image} adopts convolutional sparse representation for image fusion, extracts multi-layer features, and utilizes these features to generate fusion images. To eliminate the need for laborious manual design, an end-to-end fusion structure based on CNN is proposed, rendering the fusion process more flexible and straightforward. U2Fusion~\cite{xu2020u2fusion} adopts the densely connected network to generate the fusion image conditioned on source images. And the weight block is used to obtain two data-driven weights, which are used as the retention of features in different source images to measure the quality and information of the image. Furthermore, it combines continuous learning and other technologies to achieve multi-task fusion. It is the first all-in-one image fusion method. In addition, in recent years, the image fusion with high-level visual task has made great progress~\cite{liu2022target}. Recently, the diffusion-based image fusion came into people's view. DDFM~\cite{zhao2023ddfm} utilizes the bayesian theory, score-matching and pretrained diffusion model to get the awesome results.

\begin{figure*}[htbp]
\centering
\includegraphics[width=0.98\linewidth]{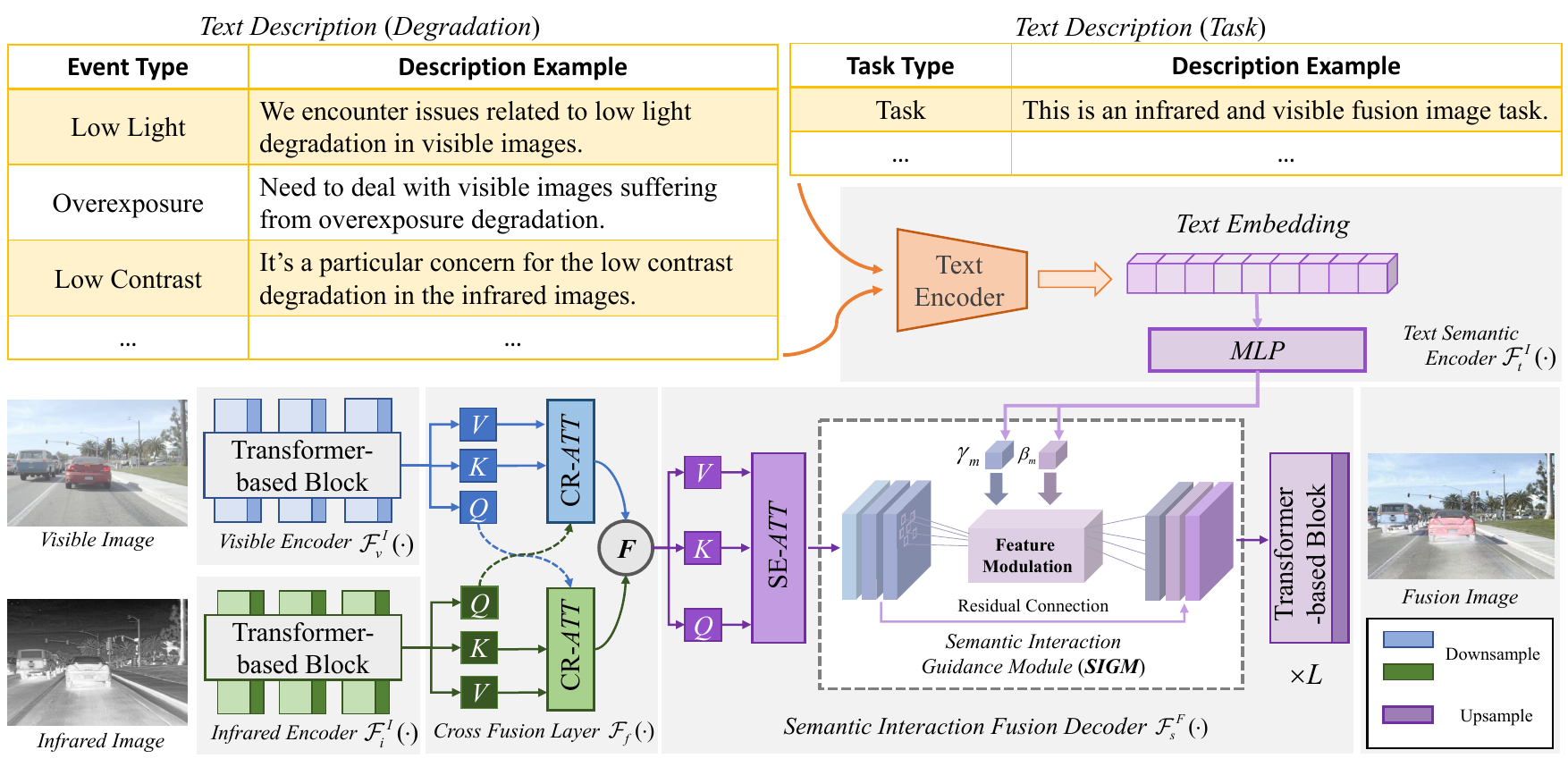}
\vspace{-0.1in}
\caption{The workflow of Text-IF. It contains two important parts, which are the image fusion pipeline and the text semantic feature encoder. Text semantic features are used to guide image fusion through the Semantic Interaction Guidance Module (\textbf{SIGM}).}
\label{fig:2}
\end{figure*}

\textbf{Text-Image Models.}
With advancements in Transformers and representation learning, coupled with the support of large datasets, multi-modal text guided image models have achieved success. CLIP~\cite{radford2021learning} is built upon two neural network-based encoders that use a contrastive loss for aligning pairs of image and text. Owing to extensive data and unsupervised training, it has the powerful zero-shot recognition and the robust text, image feature extraction capabilities. Numerous methods for text-driven image generation and processing have been proposed with the support of the CLIP model. Style-CLIP~\cite{patashnik2021styleclip} designs a text guided interface to the StyleGAN~\cite{karras2020analyzing}, allowing the alteration of real images using text prompts. In addition to GAN models, the diffusion model with text conditions also catch a lot of attentions. DiffusionCLIP~\cite{kim2022diffusionclip} proposes the diffusion model with the CLIP for text-driven image processing. Besides, stable diffusion~\cite{rombach2022high} combines the diffusion model with the text encoder and the attention mechanism to achieve the effect of text controlling image generation. Through text guidance, it can customize the effect of image generation, image processing and other tasks, and realize interactive multi-modal fusion control.

Existing image fusion methods are helpless in complex scene facing degradations. It is troublesome and not well-done even if equipped with the SOTA image restoration models. Furthermore, it is difficult to achieve interactive high-quality image fusion for users without professional knowledge. Thus, it is necessary to innovatively introduce the text guided image fusion framework for simply using.

\section{The Proposed Method}
This section describes the workflow of Text-IF, as shown in Fig.~\ref{fig:2}. We introduce from the perspective of the image fusion pipeline and the text interaction guidance architecture, including the text semantic encoder and the semantic interaction guidance module.

\subsection{Problem Formulation}
General image fusion methods formulate the image fusion task into taking two source images (\textit{e.g.}, $I_{vis}, I_{ir}$) as the input and a fusion network (\textit{e.g.}, $\theta_{n}$) to obtain an immobilized image fusion result. The network is designed to learn the mapping predefined fusion function $\mathcal{F}_{if}$ corresponding to the fusion task. In simple terms, it can be described as:
\begin{equation}
\label{eq1}
I^{f}=\mathcal{F}_{if}(I_{vis}, I_{ir}; \theta_{n}).
\end{equation}

It means that the fusion network tends to learn a relatively fixed fusion strategy. Moreover, in complex environments, such as the source images suffering from degradation, this kind of task paradigm is helpless. We study leveraging the text for breaking the traditional single fusion result along with difficulty in the quality improvement on degradations, and explore the novel text guided image fusion paradigm. Due to the introduction of text semantics, this fusion task is rewritten as:
\begin{equation}
\label{eq2}
I^{f}=\mathcal{F}_{s-if}(I_{vis}, I_{ir}, T_{text}; \theta_{n-s}).
\end{equation}

The origin mapping fusion function $\mathcal{F}_{if}$ is extended to $\mathcal{F}_{s-if}$ with text semantic information guidance. Through the interaction of text semantics $T_{text}$, the image fusion network $\theta_{n-s}$ can achieve a more customized and flexible fusion effect according to the text given by the users. At the same time, it can also restore and fuse images freely in the face of various source image degradation.

\subsection{Image Fusion Pipeline}

\textbf{Image Encoder}. The image encoder takes the source visible and infrared images as the input, respectively. Considering the spatial and deep information extraction, to obtain a comprehensive and accurate representation, we adopted Transformer/Restormer\cite{zamir2022restormer}-based blocks as the base feature extractor. In simple terms, it can be stated as follows:
\begin{equation}
\label{eq3}
F_{vis}=\mathcal{F}^{I}_{v}(I_{vis}), F_{ir}=\mathcal{F}^{I}_{i}(I_{ir}),
\end{equation}
where $I_{vis}\in \mathbb{R}^{H\times W\times 3}$ and $I_{ir}\in \mathbb{R}^{H\times W\times 1}$ represent the visible and infrared images. $H,W$ denote the height and width of the image. $\mathcal{F}^{I}_{v}$ and $\mathcal{F}^{I}_{i}$ are the visible image and infrared image encoder, respectively.

\textbf{Cross Fusion Layer}. The cross fusion layer aims to integrate the feature information from different modalities. In order to comprehensively integrate the features across all dimensions, the cross-attention (CR-ATT) is firstly used to interact the features of different modalities. Specifically, it can be expressed as follows:
\begin{equation}
\label{eq3}
\footnotesize
\{Q_{v}, K_{v}, V_{v}\} = \mathcal{F}^{v}_{qkv}(F_{vis}), \quad \{Q_{i}, K_{i}, V_{i}\}= \mathcal{F}^{i}_{qkv}(F_{ir}),
\end{equation}
where $F_{vis},F_{ir}$ denote features from the visible encoder and infrared encoder. Subsequently, we exchange the queries $Q$ of two modalities for spatial interaction:
\begin{equation}
\label{eq3}
\footnotesize
F_{f}^{i}=softmax(\frac{Q_{v}K_{i}}{d_{k}})V_{i}, \quad
F_{f}^{v}=softmax(\frac{Q_{i}K_{v}}{d_{k}})V_{v},
\end{equation}
where $d_{k}$ is the scaling factor. Finally, we concatenate the results obtained by the cross-attention calculation through $F_{f}^{0}=Concat({F_{f}^{i},F_{f}^{v}})$ to get the fusion features.

\textbf{Semantic Interaction Fusion Decoder}. The features of the cross fusion layer output are firstly enhanced by self-attention
 (SE-ATT), \textit{i.e.}, $\hat{F}^{0}_{f}=softmax(Q_{f}K_{f}/d_{k})V_{f}$. $Q_{f}$, $K_{f}$ and $V_{f}$ are the $Q$, $K$, and $V$ of $F^{0}_{f}$. Subsequently, it is interactively guided by semantic text features.

The semantic interaction fusion decoder is designed to interact text semantic features $F_{text}\in \mathbb{R}^{N\times L}$ and image fusion features $F_{f}$.  Specifically, it is constructed by the Transformer-based decoder block and Semantic Interaction Guidance Module (SIGM) which will be introduced in Sec.~\ref{TIGA}. The fusion decoder block and SIGM are tightly coupled together in a multi-stage cascade to achieve the effect of dense regulation and guidance. Briefly, the semantic interaction fusion decoder can be described as:
\begin{equation}
\label{eq4}
F_{f}^{k+1}=\{\mathcal{F}_{f}^{D}(\mathcal{L}_{f}^{s}(F_{f}^{k}, F_{text}))\}_{r},
\end{equation}
where $F_{f}^{k}$ denotes the image fusion feature at the $k$-th block stage. $\{\cdot \}_{r}$ represents the multilevel repetition. $\mathcal{F}_{f}^{D}$ and $\mathcal{L}_{f}^{s}$ denote the Transformer-based block and SIGM. Note that the upsampling is required between the levels of decoders to correspond to the downsampling at the encoder.

\subsection{Text Interaction Guidance Architecture}
\label{TIGA}
The preset image fusion pipeline can effectively obtain the corresponding fusion features $F_{f}$. And the Text Interaction Guidance Architecture is the key part to couple the text semantic information and image fusion. 

\textbf{Text Semantic Encoder}. Given a text $T_{text}$ that provides the corresponding semantic feature to guide the image fusion network to obtain the specified fusion result (\textit{e.g.}, specify the task type and the degradation type), the text semantic encoder of the text interaction guidance architecture should transfer it into the text embedding. As a large pre-trained visual language model, CLIP has a good effect on text feature extraction. We tend to freeze the good text encoder from the CLIP to maintain good linguistic consistency. With $\{\cdot\}_{e}$ denoting the frozen weights, this process can be expressed as:
\begin{equation}
\label{eq4}
F_{text}=\{\mathcal{F}^{I}_{t}\}_{e}(T_{text}),
\end{equation}
where $F_{text}\in \mathbb{R}^{N\times L}$ denotes the text semantic feature. In different but semantically similar texts, the extracted features should be close in the reduced Euclidean space.

Furthermore, we design the MLP $\Phi_{m}^{i}$ to mine this connection and further map the text semantic information and the semantic parameters. Therefore, it can be obtained:
\begin{equation}
\label{eq5}
\gamma_{m} = \Phi_{m}^{\uppercase\expandafter{\romannumeral1}}(F_{text}), \ \beta_{m} = \Phi_{m}^{\uppercase\expandafter{\romannumeral2}}(F_{text}),
\end{equation}
where $\Phi_{m}^{\uppercase\expandafter{\romannumeral1}}$ and $\Phi_{m}^{\uppercase\expandafter{\romannumeral2}}$ are the chunk operations of $\Phi_{m}$ to form the semantic parameters.

\textbf{Semantic Interaction Guidance Module (SIGM)}. In the semantic interaction guidance module, semantic parameters interact through feature modulation and fusion features $F_{f}^{i}$, so as to obtain the effect of guidance. The feature modulation consists of scale scaling and bias control, which adjust the features from two perspectives, respectively. In particular, a residual connection is used to reduce the difficulty of network fitting. For simplicity, it can be described as:
\begin{equation}
\label{eq6}
\hat{F}_{f}^{i} = (1 + \gamma_{m}) \odot F_{f}^{i} + \beta_{m},
\end{equation}
where $\odot$ denotes Hadamard product. $F_{f}^{i}$ denotes the fusion feature.  $\hat{F}_{f}^{i}$ is that with textual semantic information.

\subsection{Loss Functions}
\label{loss functions}
The loss function largely determines the type of extracted source information and the proportion relationship between source information. From the perspective of text guidance, we not only hope to solve various degradation problems through text freedom. It is also expected that the text can autonomously choose the optimal loss corresponding to the fusion task according to the needs of the users. Therefore, in the text guided image fusion task, the construction of loss function is a relation of open-set multi-point mapping.

The fusion-related losses include the intensity loss, structural similarity (\textit{SSIM}) loss~\cite{zhao2016loss}, maximum gradient loss, and color consistency loss. Considering degradations, we adopt manually obtained high-quality visible image $I_{vis}^{g}$ and infrared image $I_{ir}^{g}$ as the constraints in the loss.

\textbf{Intensity Loss}. To highlight the salient objects in infrared and visible images, the intensity values of results are maximized to ensure the target saliency. It is defined as:
\begin{equation}
\label{eq6}
L_{int}=\frac{1}{HW}\|I_{f}-max(I_{vis}^{g},I_{ir}^{g})\|_{1}.
\end{equation}

\textbf{Structural Similarity Loss}. The structural similarity loss measures the similarity between the fusion image and the source images, so that the fusion image is similar to the source images in structure. It is expressed as:
\begin{equation}
\label{eq6}
\footnotesize
L_{SSIM}(t)\!=\!\left(1\!-\!SSIM(I_{f}, I_{vis}^{g})\right)\!+\!\delta_{ir}(t)\left(1\!-\!SSIM(I_{f}, I_{ir}^{g})\right),
\end{equation}
where $\delta_{ir}(t)$ denotes the ratio of infrared structural similarity loss which is a function of the text semantic.

\textbf{Maximum Gradient Loss}. This loss preserves the maximum edges in two source images. Then, a clearer texture representation can be obtained. It can be expressed as:
\begin{equation}
\label{eq6}
L_{grad}=\frac{1}{HW}\|\nabla I_{f}-max(\nabla I_{vis}^{g}, \nabla I_{ir}^{g})\|_{1}.
\end{equation}

\textbf{Color Consistency Loss}. It keeps the fusion image and the visible image with consistent colors. We transpose the image to YCbCr space and constrain it with the Euclidean distance of Cb and Cr channels. It can be expressed as:
\begin{equation}
\label{eq6}
L_{color}=\frac{1}{HW}\|\mathcal{F}_{CbCr}(I_{f})- \mathcal{F}_{CbCr}(I_{vis}^{g})\|_{1},
\end{equation}
where $\mathcal{F}_{CbCr}$ denotes the transfer function of RGB to CbCr.

\textbf{Total Loss}. The overall loss function is a combination of fusion-related losses and is regulated by semantic information. Simply, it can be expressed as:
\begin{equation}
\begin{aligned}
\label{eq6}
L_{total}=&\alpha_{int}(t)L_{int}+\alpha_{SSIM}(t)L_{SSIM}(t)+\\ &\alpha_{grad}(t)L_{grad}+\alpha_{color}(t)L_{color},
\end{aligned}
\end{equation}
where $\alpha_{int}(t)$, $\alpha_{SSIM}(t)$, $\alpha_{grad}(t)$, and $\alpha_{color}(t)$ are semantically regulated hyper-parameters related to the task $t$. The trade-off of the fusion result plays a large role.

\section{Experiments}
In this section, we first introduce the implementation details and relevant configuration. Then, the effectiveness and superiority of the proposed method are evaluated through qualitative and quantitative comparisons. In particular, the specific results of text guided image fusion are analyzed. Finally, ablation experiments are performed.

\subsection{Implementation Details and Datasets}
\textbf{Implementation Details.} The proposed Text-IF is trained with the text guided image fusion data. The learning rate is 0.0001 with the AdamW optimizer. And the batch size is 16. The source images are cropped to $96 \times 96$. The set of hyper-parameters $\{\alpha_{int}(t), \alpha_{SSIM}(t), \alpha_{grad}(t), \alpha_{color}(t)\}$ is essentially a discrete complex map associated with the semantic text. See the additional material for details. All the experiments are conducted on the NVIDIA GeForce RTX 3090 GPU with PyTorch framework.

\textbf{Datasets.} To verify generalization, the commonly used infrared and visible image fusion datasets are MSRS~\cite{tang2022piafusion}, MFNet~\cite{ha2017mfnet}, RoadScene~\cite{xu2020u2fusion} and LLVIP~\cite{jia2021llvip}. These original datasets come with degradations in different situations, such as low light, overexposure, \textit{etc.}, in visible images, and low contrast, noise, \textit{etc.}, in infrared images. We select images where the scene is different and use manual restoration to obtain the high-quality source image, and add the corresponding hundreds of description instructions to ensure that users can input text freely for interaction. Totally we use 3618 image pairs for training, and 1135 for testing.

\textbf{Metric.} Metrics include the sum of the correlations of differences (SCD)~\cite{aslantas2015new}, standard deviation (SD), information entropy (EN)~\cite{ma2019infrared}, visual information fidelity (VIF)~\cite{han2013new}, quality of gradient-based fusion ($Q^{AB/F}$)~\cite{ma2019infrared}, CLIP-IQA~\cite{wang2023exploring}, NIQE~\cite{mittal2012making}, MUSIQ~\cite{ke2021musiq}, BRISQUE~\cite{mittal2012no}, and spatial frequency (SF)~\cite{eskicioglu1995image}. Higher values of SCD, SD, EN, VIF, $Q^{AB/F}$, CLIP-IQA, MUSIQ, and SF indicate higher quality of the fusion image. Besides, the lower values of NIQE, and BRISQUE indicate the higher quality. 

\textbf{SOTA Competitors.} We compare the proposed method with several state-of-the-art methods on multiple datasets. The methods for comparison include UMF-CMGR~\cite{wang2022unsupervised}, TarDAL~\cite{liu2022target}, ReCoNet~\cite{huang2022reconet}, MURF~\cite{xu2023murf}, U2Fusion~\cite{xu2020u2fusion}, MetaFusion~\cite{zhao2023metafusion}, and DDFM~\cite{zhao2023ddfm}.

\subsection{Comparison without Text Guidance }
Existing image fusion methods do not have semantic guidance. For comparison fairness, we first merely compare the fusion performances where no semantic guidance is provided. At this point, Text-IF uses default text. It means that no additional semantic information is introduced.

\begin{figure*}[!ht]
\centering
\includegraphics[width=0.98\linewidth]{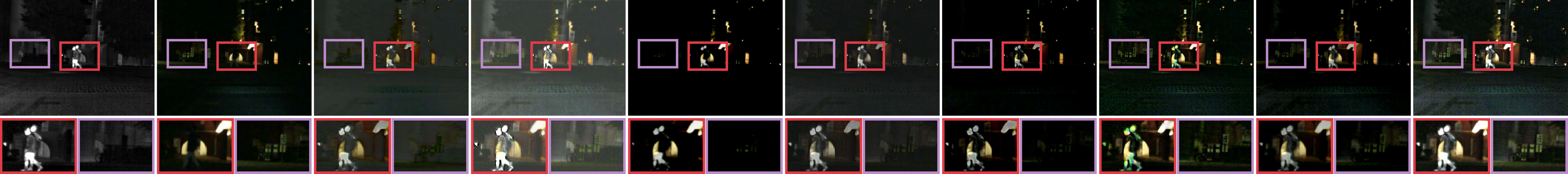}
\includegraphics[width=0.98\linewidth]{images/3-2.pdf}
\includegraphics[width=0.98\linewidth]{images/3-3.pdf}
\includegraphics[width=0.98\linewidth]{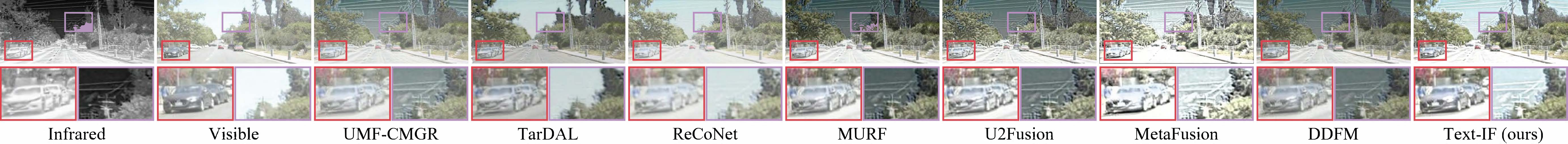}
\vspace{-0.11in}
\caption{Qualitative comparison of our Text-IF without text guidance (without additional semantic information) and existing image fusion methods. From top to bottom: data from MSRS, two groups of data from LLVIP, and data from RoadScene datasets, respectively.}
\label{default_qual}
\vspace{-0.08in}
\end{figure*}

\begin{table*}[!t]
    \renewcommand{\arraystretch}{1.15}
    \centering
    \caption{Quantitative comparison of our Text-IF without text guidance (without introducing additional semantic information) and existing image fusion methods on the MSRS, LLVIP, and RoadScene datasets (\textbf{Bold}: optimal performance).}
    \vspace{-0.1in}
    \resizebox{0.95\textwidth}{!}{
        \begin{tabular}{c@{\,~~}|c@{\,~~~}c@{\,~~~}c@{\,~~~}c@{\,~~~}c@{\,~~~}
        |c@{\,~~~}c@{\,~~~}c@{\,~~~}c@{\,~~~}c@{\,~~~}|
        c@{\,~~~}c@{\,~~~}c@{\,~~~}c@{\,~~~}c@{\,~~~}}
            \toprule
             \multicolumn{1}{c|}{\multirow{2}{*}{\textbf{Methods}}}& \multicolumn{5}{c|}{\textbf{MSRS Dataset}} & \multicolumn{5}{c|}{\textbf{LLVIP Dataset}} & \multicolumn{5}{c}{\textbf{RoadScene Dataset}} \\\cline{2-16}
              & SCD & SD & EN & VIF & $Q^{AB/F}$ &  SCD & SD & EN & VIF & $Q^{AB/F}$ &  SCD & SD & EN & VIF & $Q^{AB/F}$\\
            \midrule
            UMF-CMGR  & 0.981 &20.819 &5.600 &0.430 &0.266 &1.029 &31.501 &6.569	&0.509 &0.352 &1.613 &36.251 &6.973 &0.554 &0.429\\
            TarDAL  & 1.484 &35.460 &6.347 &0.673 &0.426 &0.817 &39.070 &5.349	&0.330 &0.252 &1.415 &42.609 &7.054 &0.525 &0.391\\
            ReCoNet  & 1.191 &44.374 &3.895 &0.438 &0.367 &1.345 &41.234 &5.514	&0.513 &0.364 &1.589 &37.580 &6.822 &0.504 &0.354\\
            MURF  & 0.868 &16.431 &5.047 &0.413 &0.327 &0.514 &21.834 &6.051	&0.386 &0.206 &1.576 &36.788 &6.992 &0.484 &0.432\\
            U2Fusion   & 1.182 &23.541 &5.246 &0.506 &0.372 &0.757 &23.614 &5.972	&0.552 &0.341 &1.498 &30.969 &6.739 &0.513 &0.467\\
            MetaFusion & 1.486 &39.432 &6.368 &0.726 &0.478 &1.317 &42.446 &6.823	&0.833 &0.493 &1.581 &\textbf{50.613} &7.223 &0.512 &0.338\\
            DDFM   & 1.550 &32.749 &5.693 &0.622 &0.431 &1.414 &38.346 &6.979	&0.549 &0.220 &\textbf{1.864} &44.925 &7.226 &0.544 &0.413\\
            \textbf{Text-IF (ours)}   & \textbf{1.681} &\textbf{44.564} &\textbf{6.789} &\textbf{1.046} &\textbf{0.676} &\textbf{1.591}	&\textbf{48.834} &\textbf{7.325} &\textbf{1.011} &\textbf{0.616} &1.572 &48.962 &\textbf{7.332} &\textbf{0.739} &\textbf{0.578}\\
            \bottomrule
        \end{tabular}
    }\label{quan_default}
    \vspace{-0.09in}
\end{table*}

\textbf{Qualitative Comparison.} The results on three datasets are shown in Fig.~\ref{default_qual}. Text-IF shows three distinctive advantages thanks to the Transformer-based pipeline with high expressive power and the implicit embedding image restoration prior. First, our results can highlight the thermal targets. As shown in the first three groups of results, the pixel intensity of thermal targets in our results are the highest. It indicates that the thermal targets in our results are the most prominent. Second, our results exhibit more appropriate brightness and provide more details. In the second and third groups, most regions of our results show higher pixel intensity than the results of competitors. In this case, more scene content can be presented clearly. Last, our results can present more vibrant and natural colors. As shown in the last example, in our result, the colors of cars and trees are more similar to those of visible images. By reducing the interference of infrared images on the color information in visible images, our fusion results are more conducive to visual perception from the perspective of colors.

\textbf{Quantitative Comparison.} The quantitative results tested with five metrics on three datasets are reported in Tab.~\ref{quan_default}. On the MSRS and LLVIP datasets, our method performs best on all the five metrics, especially showing significant advantages in SCD and VIF. On the RoadScene dataset, our method also performs optimally on three metrics. The results on EN, VIF, and $Q^{AB/F}$ reflect that even without text guidance, our method can also generate fusion results with the most information, cause the least distortion between the fusion and source images, and transfer the most edges into the fusion image. The optimal or comparable results on SCD and SD reflect that our results show little fusion distortion and high contrast (good visual effect). From the perspective of metrics, the superiority on multiple metrics indicates the comprehensiveness of the proposed method in terms of fusion performance. From the perspective of datasets, the superiority of the proposed method on multiple datasets reflects its generalization in multiple data distributions and multiple types of scenarios.

\subsection{Comparison with Text Guidance}
In real scenarios, source images may usually suffer from various degradations, {\textit{e.g.}}, poor illumination, noise, and low contrast. Existing image fusion methods cannot handle these degradations, resulting in unsatisfactory fusion results while our method can handle them through simple text guidance. Thus, for fairness, we combine existing image fusion methods with image restoration methods for comparison. SOTA image restoration models for different degradations include URetinex~\cite{wu2022uretinex} for low-light image enhancement, AirNet~\cite{li2022all} for contrast enhancement, GDID~\cite{chen2023masked} for denoising, and LMPEC~\cite{afifi2021learning} for overexposure correction. It is also worth noting that our approach uses the same model parameters in all scenarios, \textit{i.e.}, for all degradations.

\begin{figure*}[!t]
\centering
\includegraphics[width=0.99\linewidth]{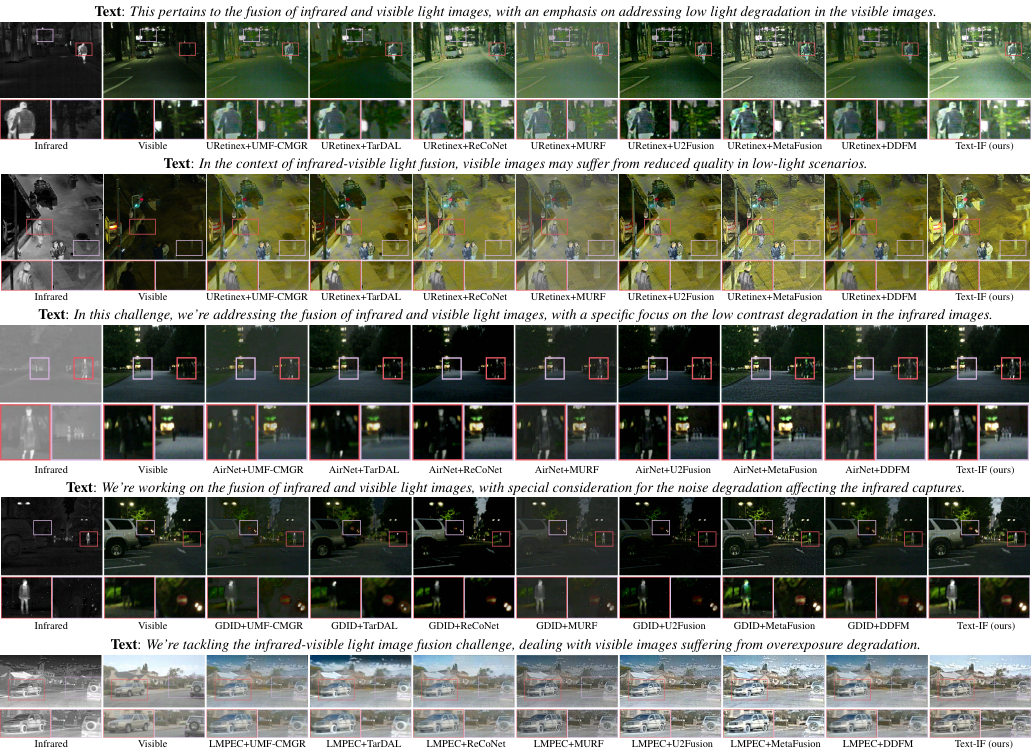}\\
\vspace{-0.12in}
\caption{Comparison of our Text-IF with semantic text guidance and the combination of existing image restoration and fusion methods on degraded source images. The semantic text is reported above each group of images. Degradations from top to bottom: low-light visible (MSRS), low-light visible (LLVIP), low-contrast infrared (MFNet), noised infrared (DN-MSRS), over-exposed visible (RoadScene).}
\label{semantic_text_qual}
\vspace{-0.13in}
\end{figure*}

\textbf{Qualitative Comparison.}
The qualitative results of Text-IF and the results of combining SOTA image restoration and image fusion competitors on degraded source images are shown in Fig.~\ref{semantic_text_qual}. In general, unlike existing methods that require manual priors for adding restoration pre-processing to fusion, Text-IF only needs to provide a simple request/description of scene and can then handle degraded source images. It avoids the tedious task of finding and switching between different restoration methods in the process of combating degradation.

Then, we compare the fusion results in various degradation scenarios in detail. First, in the first two examples, the visible images suffer from low illumination. The competitors can brighten the visible images with URetinex to some extent. However, after fusion, the low pixel intensity of infrared images still reduces the brightness of their results, and also reduces the color saturation. In comparison, our method results in more suitable brightness and brighter colors. In the third and last examples, the infrared image is of low contrast or the visible image is overexposed. In these conditions, our method can expand the dynamic range of fusion results and obtain fusion results with higher contrast and ensure the correctness of its color information at the same time. Then, the results can exhibit more clear details. In the fourth example, the infrared image suffers from obvious noise. GDID fails to remove all the noise, resulting in residual noise in fusion results. By comparison, our result shows less noise pollution, presenting higher image quality. Moreover, the prominence of thermal targets in our result is also advantageous.

\begin{table*}[!t]
    \renewcommand{\arraystretch}{1.13}
    \centering
    \caption{Quantitative comparison of our Text-IF with text guidance and the combination of existing image restoration (\textit{eir.}) and fusion methods on source images with various types of degradations (MSRS and LLVIP datasets: low-light visible images; MFNet: low-contrast infrared images; DN-MSRS: noised infrared images; RoadScene: over-exposed visible images). (\textbf{Bold}: optimal performance)}
\vspace{-0.07in}
    \resizebox{0.99\textwidth}{!}{
        \begin{tabular}{c@{\,~~}|c@{\,~~~}c@{\,~~~}c@{\,~~~}|c@{\,~~~}c@{\,~~~}
        c@{\,~~~}|c@{\,~~~}c@{\,~~~}c@{\,~~~}|c@{\,~~~}
        c@{\,~~~}c@{\,~~~}|c@{\,~~~}c@{\,~~~}c@{\,~~~}}
            \toprule
             \multicolumn{1}{c|}{\multirow{2}{*}{\textbf{Methods}}}& \multicolumn{3}{c|}{\textbf{MSRS Dataset}} & \multicolumn{3}{c|}{\textbf{LLVIP Dataset}} & \multicolumn{3}{c|}{\textbf{MFNet Dataset}} & \multicolumn{3}{c|}{\textbf{DN-MSRS Dataset}} & \multicolumn{3}{c}{\textbf{RoadScene Dataset}}\\ \cline{2-16}
             & CLIP-IQA & EN & NIQE & EN & NIQE & MUSIQ & SD & EN & MUSIQ & SD &  EN & NIQE & SF & NIQE & BRISQUE\\
            \midrule
            \textit{eir.}+UMF-CMGR  & 0.101 & 6.316 & 3.738 & 7.087 & 3.891 & 47.543 & 23.684 & 5.414 & 34.113 & 21.047 & 5.645 & 6.279 & 11.047 & 3.792 & 32.485\\
            \textit{eir.}+TarDAL  & 0.082 & 5.855 & 4.750 & 7.042 & 3.659 & 41.735 & 33.454 & 6.142 & 25.120 & 23.316 & 5.399 & 7.353 & 11.789 & 3.667 & 32.436\\
            \textit{eir.}+ReCoNet  & 0.117 & 7.216 & 5.769 & 7.109 & 4.695 & 44.187 & 41.654 & 5.161 & 29.299 & 41.525 & 4.463 & 8.631 & 10.312 & 4.785 & 37.775\\
            \textit{eir.}+MURF  & 0.111 & 5.872 & 4.199 & 6.757 & 4.177 & \textbf{50.589} & 23.741 & 5.601 & 35.626 & 20.456 & 5.280 & 6.549 & 15.605 & 3.779 & 30.594\\
            \textit{eir.}+U2Fusion   & 0.127 & 6.724 & 3.997 & 7.439 & 3.969 & 48.481 & 33.940 & 5.740 & 34.255 & 28.812 & 4.609 & 7.185 & 18.006 & 4.215 & 34.577\\
            \textit{eir.}+MetaFusion & 0.106 & \textbf{7.302} & 3.584 & \textbf{7.495} & 3.722 & 49.620 & 42.026 & 6.665 & 34.762 & 39.956 & 6.398 & 4.337 & \textbf{26.653} & 3.473 & 29.500\\
            \textit{eir.}+DDFM   & 0.094 & 6.723 & \textbf{3.465} & 7.150 & 5.184 & 35.933 & 30.465 & 6.480 & 26.902 & 27.362 & 6.120 & 4.644 & 10.493 & 3.717 & 32.334\\
            \textbf{Text-IF (ours)}   & \textbf{0.132} & 7.172 & 3.708 & 7.391 & \textbf{3.502} & 48.625 & \textbf{43.933} & \textbf{6.683} & \textbf{35.650} & \textbf{43.448} & \textbf{6.669} & \textbf{4.012} & 17.766 & \textbf{3.342} & \textbf{29.021}\\
            \bottomrule
        \end{tabular}
    }\label{text_quan}
    \vspace{-0.1in}
\end{table*}

\textbf{Quantitative Comparison.} The results on datasets in different types of degradations are reported in Tab.~\ref{text_quan}. Text-IF still achieves the overall optimal performance on all the metrics of MSRS, LLVIP, MFNet, DN-MSRS, and RoadScene datasets. The results on SD, EN and SF indicate that our method can effectively transfer the information in the fusion image. The results on CLIP-IQA, NIQE, MUSIQ  and BRISQUE show that our method can produce high quality fusion results facing the degradations.

\subsection{Performance on High-level Task}
\begin{figure}[!t]
\centering
\includegraphics[width=1\linewidth]{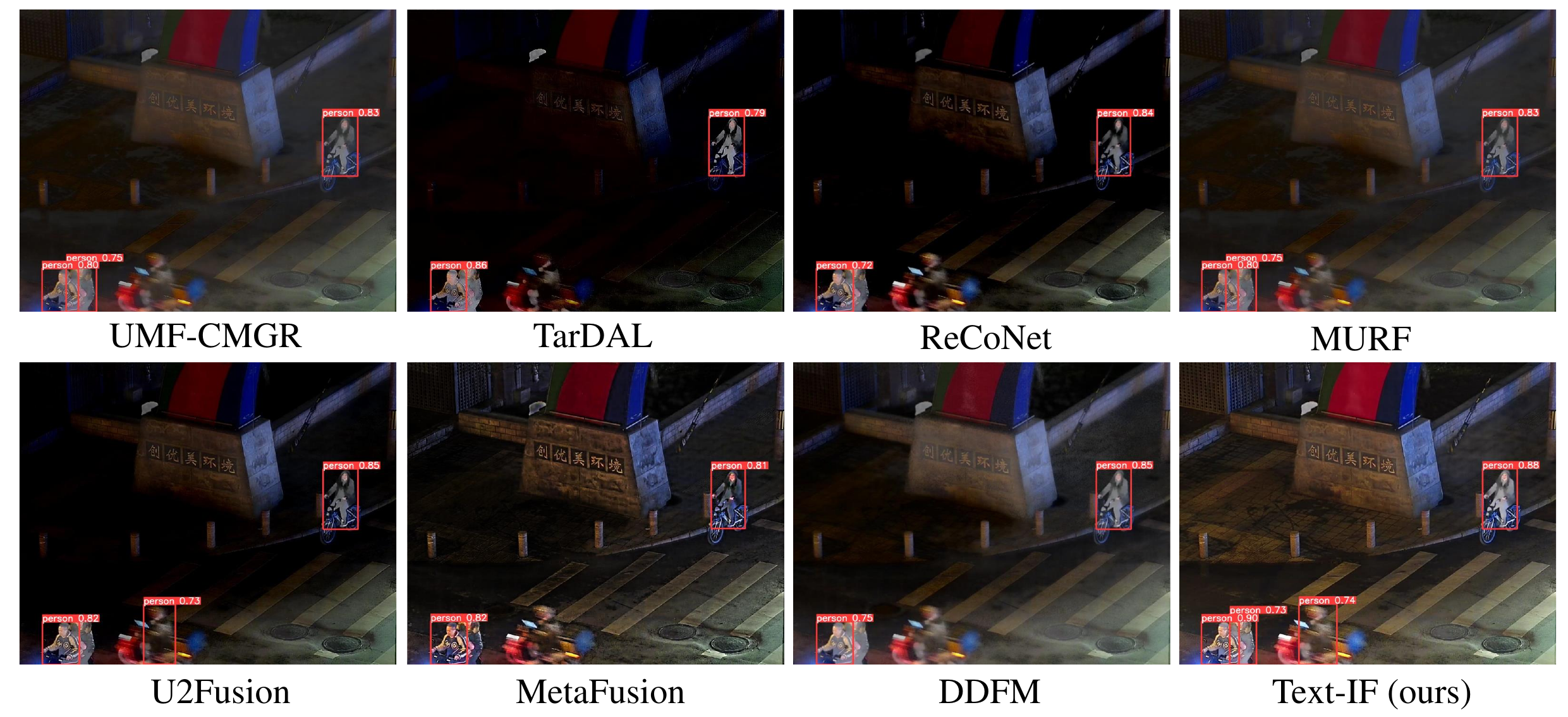}
\vspace{-0.3in}
\caption{Qualitative comparison of object detection performance on LLVIP (without introducing additional semantic information).}
\label{object_detection}
\end{figure}

To verify the performance of the fusion performance in downstream high-level vision tasks, we conduct object detection experiments on the fusion results on LLVIP dataset. We adopted YOLOv8\footnote{\url{https://github.com/ultralytics/ultralytics}} as the object detection backbone and fine-tuned it on the infrared visible light source images of LLVIP. Qualitative and quantitative experimental results are shown in Fig.~\ref{object_detection} and Tab.~\ref{object_quan}.

\begin{table}[!t]
    \renewcommand{\arraystretch}{1.12}
    \centering
    \caption{Quantitative comparison of object detection on the LLVIP dataset. Text-IF uses the default Text (without introducing additional semantic information). (\textbf{Bold}: optimal performance)}
    \vspace{-0.1in}
    \resizebox{0.49\textwidth}{!}{
        \begin{tabular}{c@{\,~~}|c@{\,~~~~}c@{\,~~~~}c@{\,~~~~}c@{\,~~~~}}
            \toprule
             \textbf{Method} & UMF-CMGR & TarDAL & ReCoNet & MURF \\
            \midrule
            mAP@0.50 & 0.925 & 0.922  & 0.916 & 0.926 \\
            mAP@0.75 & 0.659 & 0.646 & 0.617 & 0.675 \\
            mAP@0.50:0.95 & 0.599 & 0.582 & 0.568 & 0.599 \\
            \midrule
            \textbf{Method} & U2Fusion & MetaFusion	& DDFM&\textbf{Text-IF (ours)}	\\
            \midrule
            mAP@0.50 & 0.921 & 0.916 & 0.921 & \textbf{0.941}\\
            mAP@0.75 & 0.655 & \textbf{0.690} & 0.655 & 0.676\\
            mAP@0.50:0.95 & 0.591 & 0.590 & 0.592 & \textbf{0.602}\\
            \bottomrule
        \end{tabular}
    }\label{object_quan}
    \vspace{-0.23cm}
\end{table}

\textbf{Comparison with SOTA Competitors.} In terms of qualitative comparison, our proposed method Text-IF detects all the objects in the scene, while other methods have the miss detection. In terms of the quantitative comparison, Text-IF obtains the best detection performance.

\subsection{Ablation Experiment}
\begin{figure}[!t]
\centering
\includegraphics[width=0.99\linewidth]{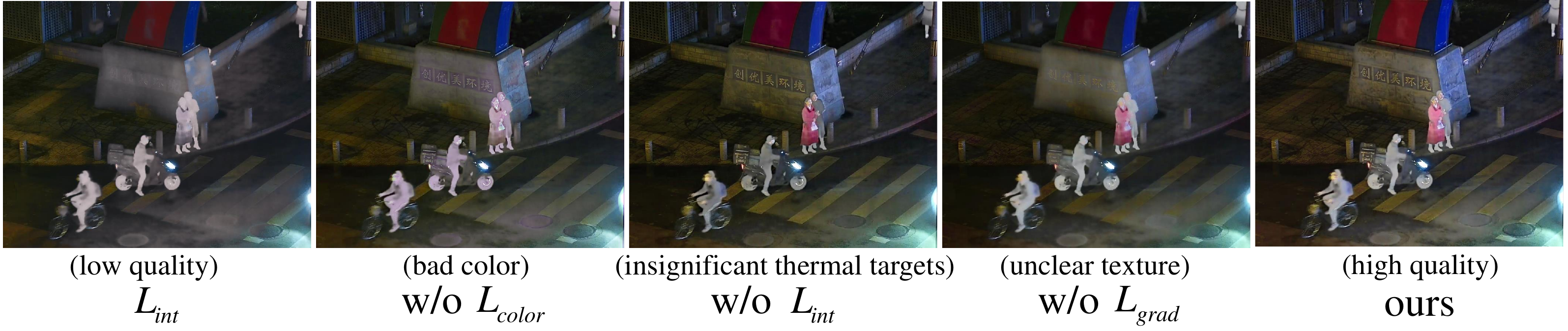}
\vspace{-0.12in}
\caption{Qualitative comparison of ablation experiment of the loss function on LLVIP.}
\label{ablation_qual}
\end{figure}

To verify the effectiveness of the proposed method, we conduct a series of ablation experiments on LLVIP dataset. It mainly includes the ablation of image fusion loss, including the intensity loss, the structural similarity (\textit{SSIM}) loss, the maximum gradient loss, and the color consistency loss. As shown in Fig.~\ref{ablation_qual} and Tab.~\ref{ablation_quan}, qualitative and quantitative results are presented.

In terms of qualitative results, the intensity loss preserves the significant thermal radiation of targets. The color loss keeps consistent color. The maximum gradient loss provides clear texture information. In terms of the quantitative results, each loss has a corresponding contribution to the final quantitative evaluation result. Our method achieves the best qualitative and quantitative evaluation among all ablation methods, which proves the effectiveness of the method.

\begin{table}[t]
    \centering
    \setlength{\tabcolsep}{10pt}
    \renewcommand{\arraystretch}{1.13}
    \caption{Quantitative comparison of the ablation experiment of the loss function on LLVIP. (\textbf{Bold}: optimal performance)}
    \vspace{-0.1in}
    \resizebox{0.48\textwidth}{!}{
    \begin{tabular}{c@{\,~~}c@{\,~~}c@{\,~~}c@{\,~~}|c@{\,~~~}c@{\,~~~}c@{\,~~~}c@{\,~~~}c@{\,~~~}}
        \toprule
        $L_{int}$ & $L_{SSIM}$ & $L_{grad}$ & $L_{color}$ & SCD & SD & EN & VIF & $Q^{AB/F}$\\
        \hline
        \checkmark &  &  &  & 1.389	 &46.147 &7.205	&0.794	&0.552 \\
                 & \checkmark &  \checkmark & \checkmark & 1.481	&42.530	&7.063	&\textbf{1.020}	&0.674\\
        \checkmark & \checkmark &  & \checkmark & 1.485 & 47.559 & 7.182 & 0.831 & 0.594 \\
        \checkmark & \checkmark & \checkmark& &  1.547	& 46.798	& 7.274	& 0.987	&\textbf{0.688} \\
        \checkmark & \checkmark & \checkmark & \checkmark & \textbf{1.591}	& \textbf{48.834}	&\textbf{7.325}	&1.011	&0.616 \\
        \bottomrule
    \end{tabular}
    }\label{ablation_quan}
    \vspace{-0.23cm}
\end{table}

\section{Conclusion}
In this paper, we extend the image fusion task and propose a novel text guided image fusion framework to address the problem that existing methods have difficulty in solving the complex scenes fusion with the degradations and getting the user-required fusion image with interactivity. Through the image fusion pipeline, the text semantic feature extraction and the semantic interaction guidance module, the goal of image fusion guided by text semantics is realized. Extensive experimental results demonstrate the obvious advantages of the proposed method in both the fusion performance and degradations treatment. It makes it possible to generate the corresponding fusion image according to the interactive user text input freely, which plays a promotive role in practice and subsequent theoretical research.

\section*{Acknowledgments}
This work was supported by NSFC (62276192).

{
    \small
    \bibliographystyle{ieeenat_fullname}
    \bibliography{main}
}


\end{document}